# A deep learning approach to using wearable seismocardiography (SCG) for diagnosing aortic valve stenosis and predicting aortic hemodynamics obtained by 4D flow MRI

Mahmoud E. Khani[1], Ethan M. I. Johnson[1], Aparna Sodhi[2], Joshua D. Robinson[1,2,3], Cynthia K. Rigsby[1,2,3], Bradly D. Allen[1], and Michael Markl[1,4]

[1]Department of Radiology, Feinberg School of Medicine, Northwestern University, Chicago, IL 60611

[2]Ann & Robert H. Lurie Children's Hospital, Chicago, IL 60611

[3]Department of Pediatrics, Feinberg School of Medicine, Northwestern University, Chicago, IL 60611

[4]Department of Biomedical Engineering, McCormick School of Engineering, Northwestern University, Evanston, IL 60208

## Abstract

In this paper, we explored the use of deep learning for the prediction of aortic flow metrics obtained using 4D flow MRI using wearable seismocardiography (SCG) devices. 4D flow MRI provides a comprehensive assessment of cardiovascular hemodynamics, but it is costly and time-consuming. We hypothesized that deep learning could be used to identify pathological changes in blood flow, such as elevated peak systolic velocity $V_{max}$ in patients with heart valve diseases, from SCG signals. We also investigated the ability of this deep learning technique to differentiate between patients diagnosed with aortic valve stenosis (AS), non-AS patients with a bicuspid aortic valve (BAV), non-AS patients with a mechanical aortic valve (MAV), and healthy subjects with a normal tricuspid aortic valve (TAV). In a study of 77 subjects who underwent same-day 4D flow MRI and SCG, we found that the $V_{max}$ values obtained using deep learning and SCGs were in good agreement with those obtained by 4D flow MRI. Additionally, subjects with TAV, BAV, MAV, and AS could be classified with ROC-AUC values of 92%, 95%, 81%, and 83%, respectively. This suggests that SCG obtained using low-cost wearable electronics may be used as a supplement to 4D flow MRI exams or as a screening tool for aortic valve disease.

## Introduction

Magnetic resonance imaging (MRI) is a crucial tool in the clinical evaluation of cardiovascular diseases. Phase contrast MRI (PC-MRI), specifically four-dimensional (4D) flow MRI, has become a routine technique for assessing the functional changes in cardiovascular blood flow in patients with heart valve, aortic, or pulmonary diseases [1-13]. 4D flow MRI provides a comprehensive evaluation of the temporal and spatial evolution of cardiovascular hemodynamics by acquiring time-resolved, three-dimensional (x-y-z) measurements of blood velocity with 3-directional velocity encoding [14-17]. Additionally, various cardiac flow metrics such as peak systolic velocity, regurgitation fraction, and wall shear stress can be retrospectively obtained from the measured blood velocities [18-29]. Despite the development of efficient 4D flow protocols for clinical applications [30-37], this technique is still considered to be costly and time-consuming due to the advanced MR sequences and computational demands of the 4D flow analysis. To address this, a preliminary evaluation of aortic flow dynamics using a cost-effective and efficient method could be valuable in diagnosing abnormalities prior to prescribing a comprehensive cardiac MRI. This study aims to investigate the utility of a wearable seismocardiography (SCG) device to predict aortic flow metrics similar to those obtained using 4D flow MRI and diagnose aortic valve pathologies.

SCG signals are non-invasive measures of cardiac vibrations obtained at the chest surface [38-52]. These vibrations are associated with heart mechanical activities such as cardiac contractions, valve closures, and changes in blood momentum [49]. For example, the atrial systole has been shown to result in a low-frequency SCG wave component, while the ventricular systole produces a high-amplitude wave, and the first and second heart sounds generate two high-frequency vibrations [43]. Additionally, simultaneous SCG and electrocardiogram (ECG) acquisition has revealed that the timing of various physiological events, such as the opening and closure of the mitral and aortic valves and their rapid filling or ejection, correspond to various features in the SCG waveforms [45]. However, SCG pulses are highly susceptible to inter-subject variations due to differences in factors such as body mass index, gender, age, and other demographic and health attributes, making it difficult to derive consistent clinical inferences [53]. Additionally, the location of the SCG device on the chest can impact the shape of the recorded waveforms. Lastly, SCG vibrations have relatively low amplitudes and are easily contaminated by subject motion artifacts or respiratory movements, which can lead to misinterpretation of diagnostic features [43].

The purpose of this study was to investigate the relationships between chest vibrations measured by SCG and cardiac flow metrics obtained using 4D flow MRI using a new deep learning approach. Our technique used ECG-synchronized scalograms of the SCG signals as input to a convolutional neural network (CNN) model for the regression or classification of cardiac parameters. In addition, we used a multi-layer perceptron (MLP) based on demographic attributes such as body mass index, gender, and age to account for inter-subject variations in SCG. We hypothesized that this deep learning approach could be used to infer pathological changes in blood flow, such as an increased peak systolic velocity ($V_{max}$) in patients with aortic valve disease, from SCG pulses. Additionally, because the presence of flow abnormalities such as elevated $V_{max}$ are indicative of aortic valve complications, we hypothesized that this technique could be used to diagnose various aortic valve conditions, such as bicuspid or mechanical aortic valves (BAV and MAV) and aortic stenosis (AS) and differentiate them from healthy subjects with normal tricuspid aortic valves (TAV).

## Methods

### 1. Study cohort

The study was approved by the Institutional Review Board (IRB) and informed consent was obtained from all participants. We recruited 46 healthy subjects (20 females, age: 45.9 ± 17.2 years) with no known history of cardiovascular disease and 31 patients with aortic valve disease (6 females, age: 32.6 ± 20.9 years). Among the patients, 1 had a normal tricuspid aortic valve (TAV), 21 had bicuspid aortic valves (BAV), and 9 had mechanical valve implants. In addition, 12 patients were diagnosed with varying degrees of aortic stenosis (AS): 2 were mild, 6 were moderate, and 4 were severe.

### 2. 4D flow MRI acquisition and analysis

4D flow MRI was performed using a spatial resolution of 1-3 mm$^3$, time resolution of 30-40 ms, and velocity encoding (venc) of 150-375 cm/s on a 1.5T or 3T scanner. The MRI was performed during free breathing with navigator respiration control and prospective or retrospective gating and covered the full volume of the thoracic aorta.

Pre-processing of the 4D flow data included correction for eddy currents, velocity anti-aliasing, and application of a deep learning tool to automatically derive a 3D segmentation of the thoracic aorta [54]. This segmentation was used to mask the 4D flow velocity data, and the ascending aorta (AAo) was manually delineated from its root to the branching vessels by placing a plane proximate to the branching vessels, perpendicular to the centerline. As shown in Fig. 1(a), peak systolic velocity $V_{max}$ in the AAo was computed from a volumetric analysis of all voxels in the AAo, with outlier-rejection [55]. All 4D flow MRI analyses were performed using a custom code in MATLAB (MathWorks, Natick, MA, USA).

### 3. SCG analysis

We used a custom-designed, wearable cardiac sensor (Fig. 1(b)) incorporating a MEMS accelerometer (1 kHz sampling rate, 2 $\mu s/m^2$ sensitivity) to acquire chest accelerations in three directions immediately before MRI. The SCG device was placed on the sternum of subjects while they were in the supine position. These acceleration recordings were beat-by-beat time-referenced to the R-waves within simultaneously acquired ECG signals. Figure 1(c) shows an example of the gating ECG signal, while Fig. 1(d-f) show the accelerations recorded in the three directions using the accelerometer over a 5 s period. At each heartbeat, one SCG pulse was calculated using the net magnitude of the accelerations along the three directions., i.e.,

$$\text{scg}_i(t) = \sqrt{\text{acc}_{i,x}^2(t) + \text{acc}_{i,y}^2(t) + \text{acc}_{i,z}^2(t)} \tag{1}$$

where subscript $i$ represents the heartbeat number, and subscripts $x$, $y$, and $z$ represent the direction of the measured accelerations. Figure 1(g) shows how the gating of SCG signals was performed using the ECG recordings. In this way, multiple SCG signals, each corresponding to one R-R period in ECG, one example shown in Fig. 1(h), were measured and processed for each subject. It should be noted that the duration of measurement was not the same for all subjects, so the number of SCG signals obtained varied for different subjects.

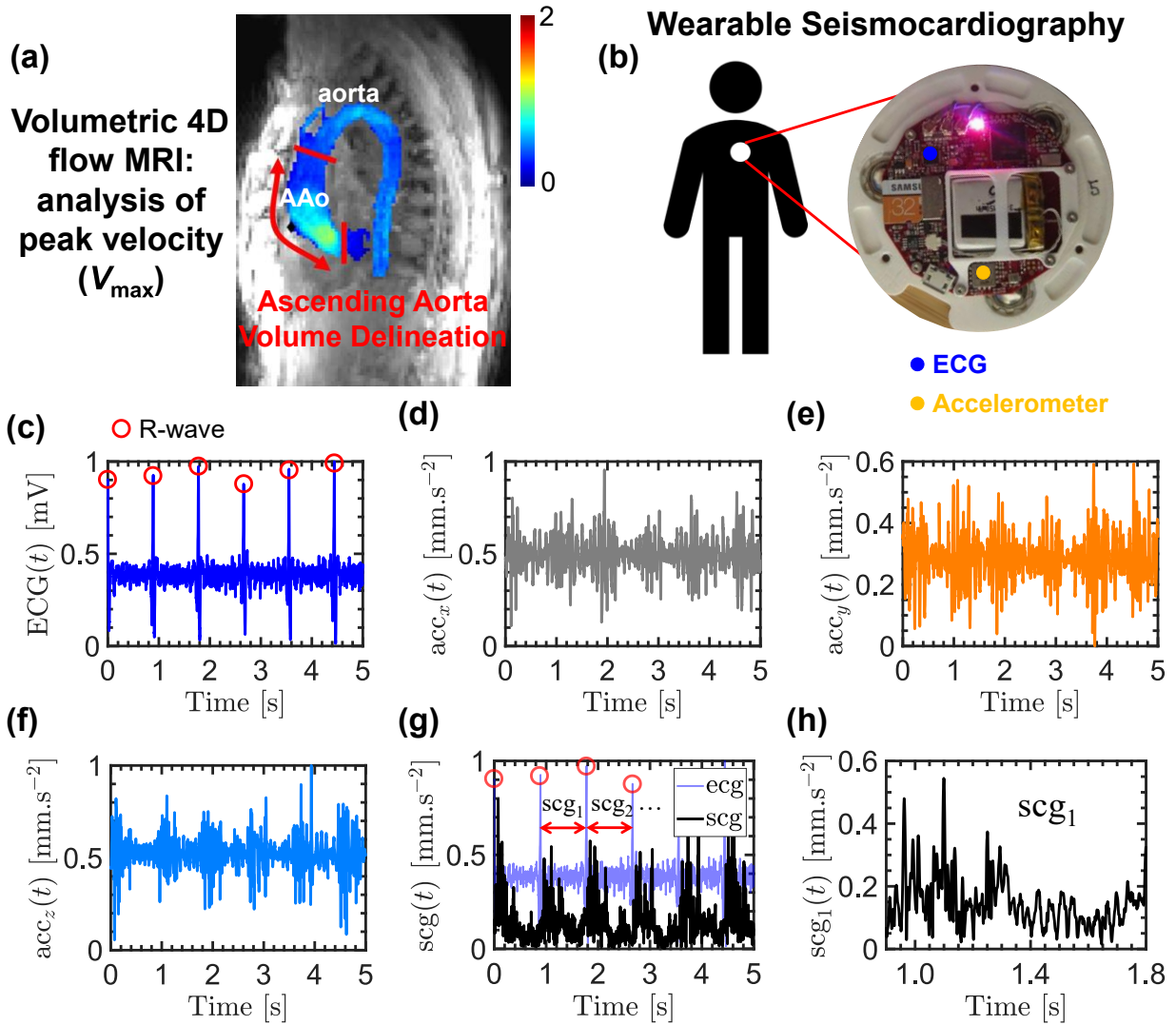

**Figure 1**. **(a)** the derivation of V$_{max}$ by AI-based 3D-segmentation of the thoracic aorta within the full 4D flow MRI measurements and manual delineation of AAo, **(b)** the SCG device, **(c)** an example of ECG signal, **(d-f)** acceleration signals measured in $x$, $y$, and $z$ directions, respectively, **(g)** gating of SCG to R-waves within simultaneously acquired ECG signals, **(h)** an example SCG pulse measured over one heartbeat.

## 4. Signal conditioning

The following steps were used to condition SCG signals in MATLAB. First, high-pass filtering was applied to remove low-frequency artifacts using a minimum-order elliptic filter with an infinite impulse response (IIR) function. The filter had a stop-band attenuation of 60 dB, a stop-band frequency of 8.4 Hz, a pass-band frequency of 10 Hz, and a pass-band ripple of 0.1. A Blackman window was used to taper the transition of boundary points into zero. Wavelet denoising was performed to improve the signal-to-noise ratio (SNR) of the signals. A symlet wavelet with four vanishing moments ('sym4') was used, and the denoising method employed was the false discovery rate (FDR) with a hard thresholding rule. The noise estimate used was level-dependent, and the number of levels of decomposition was calculated as the integer part of $\log_2 N$, where $N$ is the number of signal points. Figure 2(a) shows examples of raw and denoised SCG signals. After denoising the SCG signals, we used the Signal Quality Index (SQI) to identify and remove any outliers. The SQI is a measure of the quality of the signals and is calculated based on the distance of each SCG from a reference signal. This step helped ensure that the resulting signals were of high quality and accurately represented the subject's physiology. Detailed information about the SQI can be found in reference [56]. Briefly, for each subject, the SQI of the $i$th SCG signal, $scg_i(t)$, was calculated based on its distance from $\widehat{scg}(t)$, given by,

$$\text{SQI}_i = \exp\left(-\frac{\mathcal{D}(scg_i, \widehat{scg})}{\mathcal{L}(scg_i, \widehat{scg})}\right), \tag{2}$$

where $\widehat{scg}(t)$ represents the point-wise average of all SCGs for the subject, $\mathcal{D}(scg_i, \widehat{scg})$ represents the distance between $scg_i$ and $\widehat{scg}$, and $\mathcal{L}(scg_i, \widehat{scg})$ represents the length of $scg_i$ and $\widehat{scg}$. Following the approach described in [56], we used dynamic time warping (DTW) to calculate $\mathcal{D}(scg_i, \widehat{scg})$. The DTW algorithm was used to find the minimum Euclidean distance between two SCG signals, even if they were of different lengths. This distance, denoted as $\mathcal{L}(scg_i, \widehat{scg})$, was calculated by stretching or compressing the signals to match their lengths. A small distance in $D$ indicates a high-quality signal, while a large distance indicates a poor-quality signal.

To determine which signals were considered high quality, the SQI scores were calculated for each signal. The signals were then ranked in order of their SQI scores, and the top 95% of signals were selected as the "good" signals. Any signals with an SQI below the 5% threshold were considered outliers and were removed from further processing. This can be seen in Fig. 2(b-c), which shows examples of good SCG signals and the outliers for a single subject. The outlier signals are highly fluctuating and do not resemble a typical SCG.

## 5. Time-frequency representation – SCG scalograms

To obtain time-frequency representations of the remaining SCG signals, we used the continuous wavelet transform (CWT). We defined a CWT filter bank composed of 48 filters per octave using the analytic Morse wavelet. The filter amplitudes were normalized so that the peak magnitude of all passbands was equal to 2. The highest passband frequency was designed such that the magnitude falls to half the peak value at the Nyquist frequency.

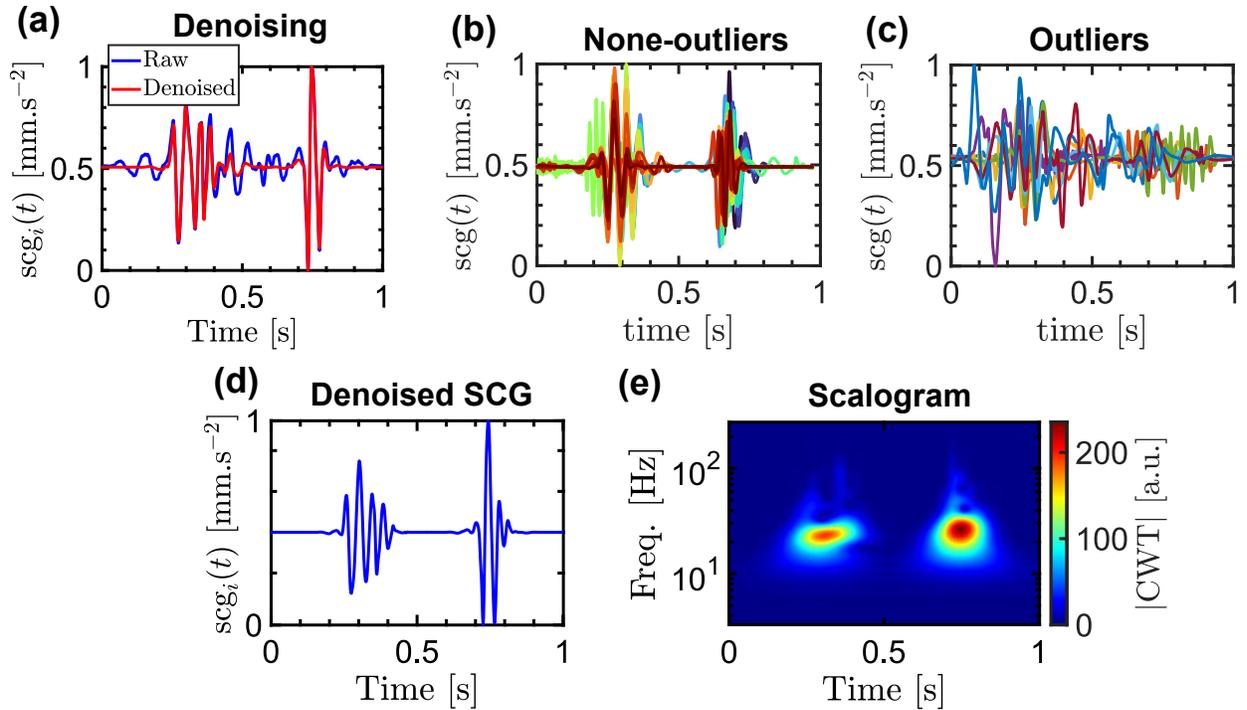

**Figure 2. (a)** comparison of a raw and denoised SCG signal, **(b-c)**, the none-outlier and outlier pulses identified based on SQI scores for a healthy subject, **(d)** an example SCG pulse composed of two main features, **(e)** the scalogram of the signal in **(d)** representing the absolute value of its CWT coefficients.

The absolute value of the CWT coefficients produced a scalogram for each SCG. Figure 2(d-e) show an example of a denoised SCG and its corresponding scalogram. It can be seen that the two main features in the SCG pulse, which occur between 0 and 0.5 seconds and between 0.5 and 1 second, appear at the same locations in the scalogram. In addition, the frequency components of these features, which lie between 10 and 100 Hz, are shown on the y-axis. These frequency values were directly estimated from their corresponding wavelet scales.

All the scalograms were converted to 256×256 images using bicubic interpolation, with each interpolated pixel representing the weighted average of the pixels in its nearest 4×4 neighborhood. This allows for easy visualization and analysis of the time-frequency characteristics of the SCG signals.

## 6. Deep learning

Figure 3(a) shows the deep learning model used for the regression of peak systolic velocity. The SCG scalograms were used as input data for a convolutional neural network (CNN) model to predict $V_{max}$ in the AAo. The CNN architecture consisted of a stack of five convolutional blocks, with each block containing a sequence of Conv2D, ReLU (rectified linear unit) activation, batch normalization, and max pooling layers. The Conv2D layer creates multiple kernels to be convolved with the input data and produces output tensors. The kernel weights and bias vectors are adjusted during backpropagation. For all convolutional blocks, the kernel size, stride, and padding parameters were set to 3×3, 1, and "same", respectively. The ReLU activation function outputs its input if it is positive and outputs zero otherwise. This adds non-linearity to the features learned by the model. The batch normalization layer normalizes each batch of data using its mean and standard deviation to facilitate backpropagation. Finally, the max pooling layer outputs the largest number within each 2×2 window of its input tensor to extract increasingly generic features by down-sampling. These layers work together to allow the CNN to efficiently learn and extract useful features from the SCG scalograms. The output of the last convolutional block was flattened and connected to two fully-connected dense units, with the first unit containing a sequence of dense, ReLU activation, batch normalization, and dropout layers. The second unit only contained dense and ReLU activation layers.

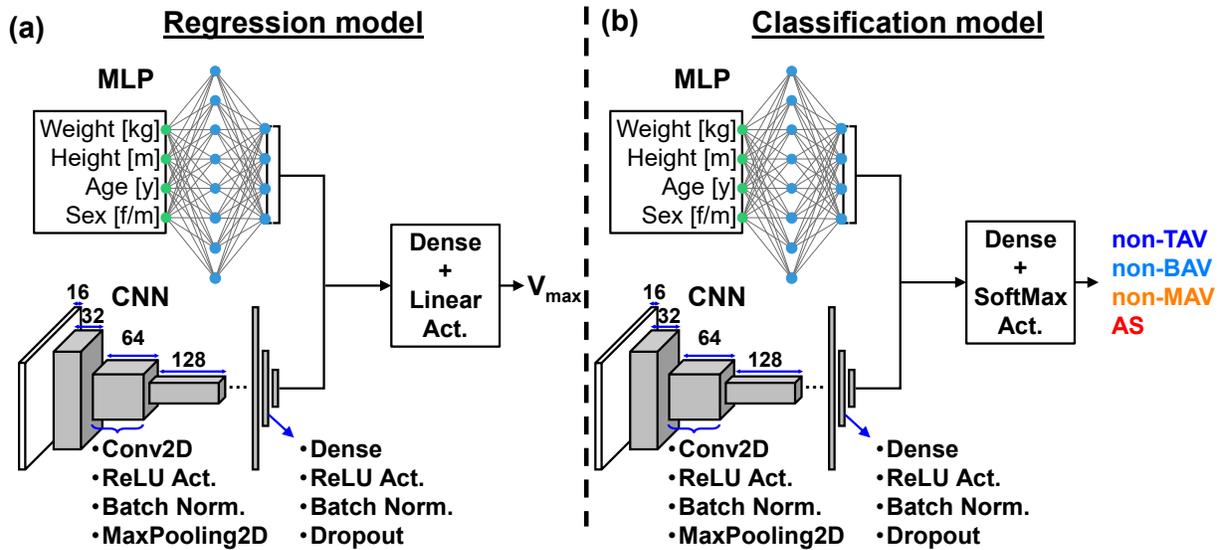

**Figure 3.** The mixture DNN model composed of a MLP and a CCN to **(a)** predict the aortic $V_{max}$ and **(b)** classify the subjects into different valve conditions: non-AS TAV, non-AS BAV, non-AS MAV, and AS, based on the demographic attributes of the subjects and scalograms of the SCG signals.

To incorporate demographic attributes of the subjects (weight, height, age, and sex) in the deep learning model, we created a multilayer perceptron (MLP) with two fully-connected dense layers. All three continuous attributes were min-max normalized, and the participants' gender was one-hot encoded (i.e. female = 1, male = 0) before being fed into the MLP.

The outputs of the MLP and CNN, which both have four nodes, were concatenated and fed into a single dense layer followed by a linear activation unit. This final layer regresses the desired flow metric, such as $V_{max}$ peak systolic velocity, using the combined information from both the demographic attributes and the time-frequency characteristics of the SCG signals.

The same model was also used to classify the subjects into different valve conditions: non-AS TAV, non-AS BAV, non-AS MAV, and AS. However, as it is shown by Fig. 3(b) for the classification task, the number of nodes in the final dense layer was equal to the number of valve condition categories, and a SoftMax activation function was used to produce the probabilities of the output belonging to each class. This allows the model to accurately classify the subjects based on their SCG signals and demographic attributes.

In both regression and classification models, the training parameters of MLP and CNN were optimized simultaneously to minimize the loss function, which was calculated as the mean percentage error between the predicted and true velocities for the regression task, and the categorical cross entropy for the classification task. The models were implemented using Tensorflow 2.6 and Keras libraries in Python 3.8 and were run on a GPU node (equipped with a 40 GB Tesla A100 GPU) within the Northwestern University Quest Computing Cluster.

To evaluate the performance of the models, we used a leave-subject-out cross-validation approach. This means that for each iteration, we used 80% of the available SCG pulses (N = 6249) for training the model and reserved the remaining 20% for testing. Importantly, the SCG pulses belonging to each subject were only included in either the training or test set, but not both, in order to avoid any potential bias in the performance metrics. We repeated this process for 10 iterations, randomly selecting different subjects to be included in the training and test sets at each trial. This allowed us to verify that the reported performance metrics were not influenced by any specific patterns in the distribution of the training and test sets. For each iteration, the model was trained for 150 epochs.

To assess the agreement between the values obtained by the DNN and 4D flow, we performed correlation and Bland-Altman analyses. The linear relationship between the two sets of values was determined using Pearson correlation coefficient, and the limits of agreement (LOA) were calculated using the Bland-Altman method. In Bland-Altman plots, each sample is represented by the mean of the two measurement techniques on the x-axis, and the difference between the two techniques on the y-axis. The mean difference is an estimate of the bias, the standard deviation (SD) of the difference measures the random fluctuations around the mean, and the LOA is defined as 1.96×SD. If the mean difference is significantly different from 0, based on a one-sample t-test (significance level $\alpha = 0.05$), this indicates the presence of a fixed bias. It is important to note that for each subject, there were multiple SCG scalograms recorded at different heartbeats. The DNN model assigned a $V_{max}$ value to each scalogram, and we used the average $V_{max}$ value over all scalograms for each subject as the representative value for that subject.

To investigate the performance of the classification model, we used receiver operating characteristic (ROC) curves. The ROC curves were constructed using the probabilities assigned by the model to each observation belonging to a particular valve condition group. An ROC curve plots the true positive rate (TPR) against the false positive rate (FPR) at different thresholds chosen from the predicted probabilities, and a higher ROC-AUC (area under the curve) indicates a better performance by the model. We also used the confusion matrix obtained by testing the classification model. In a confusion matrix, the diagonal elements represent the number of true positives ($TP$) in each category, while the off-diagonal elements at each row and column respectively show the number of false negatives ($FN$) and false positives ($FP$). In addition, we calculated and compared the precision ($\frac{TP}{TP+FP}$) and recall ($\frac{TP}{TP+FN}$) rates in diagnosing each valve condition.

# Results

### 1. Prediction of $V_{max}$

The peak systolic velocities predicted by our deep learning model were in good agreement with the velocities obtained using 4D flow MRI, as demonstrated by the low mean squared error of 0.2 m/s across ten random trials. Figure 4(a) shows the average DNN-predicted $V_{max}$ values versus the velocities obtained by 4D flow MRI for the test subjects. As mentioned earlier, we used the average $V_{max}$ value over all scalograms for each subject as the representative value. The figure shows a strong linear correlation between the estimated and measured $V_{max}$ values (y = 0.89x, r = 0.76, p $\ll$ 0.01). Additionally, Figure 4(b) shows the Bland-Altman plot for the $V_{max}$ values obtained by the DNN model and 4D flow MRI measurements. This plot indicates a low, non-significant bias (-0.08 m/s, p = 0.18) and moderate limits of agreement (±0.86 m/s).

### 2. Classification of aortic valve condition

Figure 4(c) shows the results of a classification model that was trained to identify four different types of subjects based on their SCG scalograms and demographic attributes: non-AS TAV, non-AS BAV, non-AS MAV, and AS. The model was evaluated using receiver operating characteristic (ROC) curves, which plot the true positive rate (recall) against the false positive rate. The solid lines on the ROC curves represent the mean performance of the model over 10 random iterations, while the error regions show the standard deviation of the performance across the iterations. The ROC-AUC values indicate the model's performance on each of the four classes, with values of 91.8±5.3%, 94.8±5.1%, 81.2±10.8, and 83.1±11.1% for non-AS TAV, non-AS BAV, non-AS MAV, and AS subjects, respectively.

Figure 4(d) shows a confusion matrix, which was used to evaluate the model's performance on the test set in one random trial. The rows of the matrix represent the ground-truth categories, while the columns represent the categories predicted by the model. The precision and recall rates indicate the model's ability to correctly classify subjects in each of the four classes. The precision rates for the four classes were 93%, 82%, 83%, and 49%, while the recall rates were 81%, 94%, 68%, and 86%.

# Discussion and conclusion

In this study, we evaluated the effectiveness of using deep learning and SCG scalograms to predict aortic peak systolic velocity and diagnose patients with various aortic valve complications. Our deep learning model accurately predicted peak systolic velocities, with MSE = 0.2 m/s and r = 0.76 (p<0.01), compared to velocities obtained using 4D flow MRI.

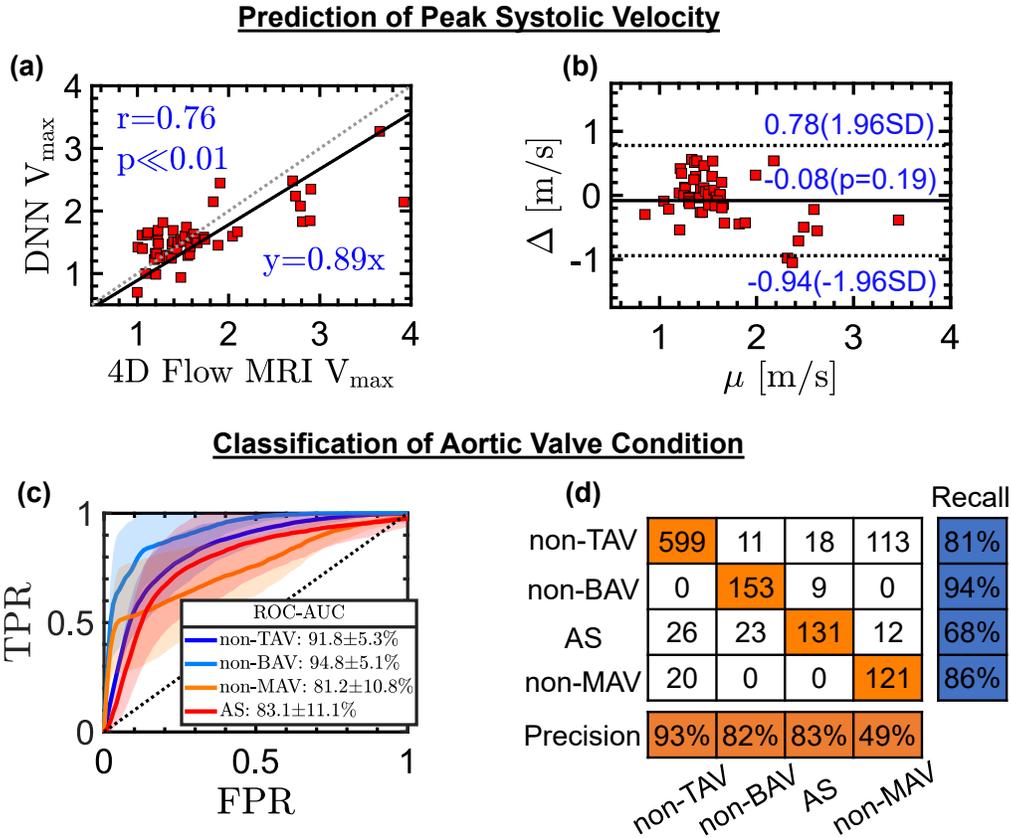

**Figure 4.** Statistical comparison between the $V_{max}$ values derived by 4D flow MRI and predicted by DNN and scalograms based on **(a)** correlation and **(b)** Bland-Altman analysis, **(c)** the ROC curves (mean±SD) for classification of non-TAV, non-BAV, non-MAV subjects, and patients with AS, **(d)** the confusion chart (at SCG level) of an iteration of training and test of the classification model.

Despite the imbalanced distribution of subjects with different types of aortic valve pathologies, we achieved high precision rates (over 82% for all categories except non-MAV) and recall rates (over 81% for all categories except AS) in the diagnosis of these conditions. Our results suggest that deep learning and SCG have significant potential as a substitute or screening tool for more advanced imaging techniques such as 4D flow MRI.

Deep learning models have been previously applied to extract various types of information from SCG signals. For example, a CNN model was developed to continuously identify R-peaks from SCG signals with high sensitivity, and it was demonstrated that heart rate variability indices obtained using this model from SCG signals were in good agreement with those obtained from ECG signals [57]. A study of 36 healthy subjects showed that deep learning could accurately map SCG signal segments to whole-body ballistocardiograms [51]. A U-Net-based cascaded framework was also proposed for estimating respiratory rate from ECG and SCG signals [58]. Other research has investigated the use of machine learning and deep learning for detecting heartbeats and heartbeat rates from SCG signals [59, 60]. In contrast to these previous works, this paper examines the correlation between wearable SCG pulses and cardiac peak systolic velocity, whose ground truth was determined using comprehensive 4D flow MRI. This study therefore expands the potential use of SCG for diagnosing cardiac abnormalities based on blood velocity, rather than just heart rate beats.

One potential application of the technique described in this work is to improve the accuracy of estimating aortic velocity for PC-MRI. A crucial parameter that needs to be specified before performing PC-MRI is the venc threshold [61]. This parameter should be set to capture the highest expected velocity within the vessel of interest while maintaining a sufficient velocity-to-noise ratio. Although the venc value is crucial for proper performance of the PC pulse sequence, it is often estimated because its optimal value is not known in advance, and sometimes a study needs to be repeated using different vencs to obtain the optimal results. The technique presented in this work could potentially be used to obtain a reliable estimation of the optimal venc value before imaging, reducing MR imaging time and costs.

A major limitation of the current study was the imbalanced distribution of subjects. In predicting peak systolic velocity, the majority of study participants were healthy, resulting in only a small percentage (about 20%) of training set samples having a velocity greater than 2 m/s. This made it more difficult for the model to accurately predict higher peak velocities in the test set, as shown in Fig. 4(a) where the correlation with ground-truth values was weaker for higher velocities compared to lower ones. In classifying valve conditions, the non-TAV group had 46 subjects, while the non-BAV, non-MAV, and AS groups each had only 10, 9, and 12 subjects, respectively. This may have contributed to the less accurate predictions of the model in diagnosing non-MAV and AS categories compared to non-TAV subjects. Furthermore, while it would have been valuable to further stratify AS severity (e.g., mild, moderate, and severe), doing so would likely result in even smaller numbers of subjects in each group. Therefore, it is necessary to expand the patient cohort to include more subjects with higher aortic velocities and different severities of AS in future studies.